\documentclass[10pt,twocolumn,letterpaper]{article}

\usepackage[pagenumbers]{arxiv}

\usepackage[linesnumbered,ruled,vlined]{algorithm2e}
\usepackage{graphicx}
\usepackage{amsmath}
\usepackage{amssymb}
\usepackage{booktabs}
\usepackage{gensymb}
\usepackage{hyphenat}
\usepackage[inline]{enumitem}
\usepackage[breaklinks,colorlinks]{hyperref}
\usepackage{multirow}

\usepackage[capitalize]{cleveref}

\begin{document}

\title{Faster Optimization-Based Meta-Learning Adaptation Phase}

\author{Kostiantyn Khabarlak \\
Dnipro University of Technology\\
Ukraine\\
{\tt\small habarlack@gmail.com}
}

\maketitle

\begin{abstract}Neural networks require a large amount of annotated data to learn. Meta-learning algorithms propose a way to decrease the number of training samples to only a few. One of the most prominent optimization-based meta-learning algorithms is Model-Agnostic Meta-Learning (MAML). However, the key procedure of adaptation to new tasks in MAML is quite slow. In this work we propose an improvement to MAML meta-learning algorithm. We introduce Lambda patterns by which we restrict which weight are updated in the network during the adaptation phase. This makes it possible to skip certain gradient computations. The fastest pattern is selected given an allowed quality degradation threshold parameter. In certain cases, quality improvement is possible by a careful pattern selection. The experiments conducted have shown that via Lambda adaptation pattern selection, it is possible to significantly improve the MAML method in the following areas: adaptation time has been decreased by a factor of 3 with minimal accuracy loss; accuracy for one-step adaptation has been substantially improved.
\end{abstract}


\blfootnote{
  This article has been published in the peer-reviewed journal. Citation: \nohyphens{K.~Khabarlak \emph{Faster Optimization-Based Meta-Learning Adaptation Phase}. Radio Electronics, Computer Science, Control, no. 1, pp. 82--92, Apr. 2022.} \href{https://doi.org/10.15588/1607-3274-2022-1-10}{DOI:~10.15588/1607-3274-2022-1-10}
}

\section*{Nomenclature}

\(N\) is a number of images per class that are given for the network training.

\(K\) is a number of classes the network is trained to distinguish between.

\(X\) is a network input, in our case, images.

\(\Phi(\theta, X)\) is a neural network.

\(\theta\) is a matrix of network weights.

\(B\) is a number of layers in the neural network.

\(\rho(\mathcal{T})\) is a distribution of all tasks.

\(\mathcal{T}_i\) is one of the tasks, consisting of Support Set \(S_i\), Query Set \(Q_i\).

\(P\) is a number of adaptation steps.

\(\theta_i^{(j)}\) is a matrix of adapted weights after \(j\) iterations that correspond to \(i\)\textsuperscript{th} task.

\(\alpha\) is an adaptation step size, \(\alpha > 0\).

\(\beta\) is a learning rate, \(\beta > 0\).

\(\Lambda\) is an adaptation pattern, which controls which neural network layers should be updated during the adaptation procedure to the current task \(\mathcal{T}\).

\section{Introduction}

The neural network accuracy for image classification has significantly improved thanks to deep convolutional neural networks. However, a very large number of images is required for such networks to train successfully. For instance, all of the ResNet~\cite{ResNet} neural network configurations from ResNet-18 to ResNet-152 (18 and 152 layers deep correspondingly) are trained on the ImageNet dataset~\cite{ImageNet}, which contains 1,281,167 images and 1,000 classes (about 1,200 samples per class). Obviously, for many of the practically significant tasks it is impossible to collect and label a dataset that large. Thus, learning deep convolutional networks from scratch might yield poor results. Because-of that, on the smaller datasets typically an approach called transfer learning is used instead. That is, an ImageNet pretrained network of a particular architecture is taken and then further finetuned on the target (smaller) dataset~\cite{ResNet,DenseNet,WideResNet}. However, training on few examples per class is still a challenge. This contrasts to how we, humans, learn, when even a single example given to a child might be enough. Also, it is hard to estimate the quality of a certain ImageNet pretrained network on the target dataset. Hence, we get a model selection problem: if the model A is better than the model B on ImageNet, will it be better on our small dataset? A promising approach to resolving both of these problems is to use meta-learning or its benchmark known as few-shot learning. Meta-learning trains the network on a set of different tasks, which are randomly sampled from the whole space of tasks. By learning the network in such a way, it is assumed that the network will learn features that are relevant to all of the tasks and not only to the single one, \ie, will learn more general features.

In this work we focus on one of the most prominent optimization-based meta-learning methods, called Model-Agnostic Meta-Learning (MAML)~\cite{MAML}. This method has become a keystone, and as it will be shown in the literature overview section, many of the newer method base on its ideas. Training of the MAML method is split into the so-called adaptation and meta-gradient update phases. It has been shown that the adaptation phase of the MAML is quite slow to perform~\cite{MetaLearningImplicitGradients}, and in general, slow neural network execution is a major problem for applications~\cite{FastFacialLandmarkSurvey}. In this work we introduce gradient update patterns, \ie, a selective update of the neural network weights during the adaptation phase.

The purpose of this work is to show that by carefully selecting the newly-proposed gradient update pattern, it is possible to increase the speed of MAML adaptation phase, and to significantly improve MAML performance in case, when only 1 adaptation phase is used. The testing results will be shown on a publicly available few-shot learning dataset CIFAR-FS~\cite{MetaLearningClosedFormSolvers}.

\section{Problem Statement}\label{sec:problem-statement}

The goal behind meta-learning is to train a neural net-work \(\Phi(\theta, \cdot)\), that is capable of adapting to the new previously unknown tasks given a small number of examples. Meta-learning is also said to be learning to learn problem. The training procedure is defined using a concept of tasks, that are sampled from the whole task space \(\rho(\mathcal{T})\) of the problem domain. The task is a tuple \(\mathcal{T}=\{S, Q\}\), consisting of the so-called Support Set \(S = \{X_S, y_S\}\) and Query Set \(Q = \{X_Q, y_Q\}\)~\cite{MAML,PrototypicalNetworks,MiniImageNetRavi2017,HowToTrainMAML}. Support Set \(\{X_S, y_S\}\) is used to adapt (or train) the network to the new task. The set \(S\) is small. \(X_S\) are the network inputs, \(y_S\) are the expected predictions. The number of examples per class is denoted as \(K\) and written as \(K\)-shot. \(K\) is typically in range from 1 to 20, although no hard upper-bound is defined. \(X_Q, y_Q\) are the query inputs and expected outputs correspondingly, which are used to evaluate the network. Number of classes \(N\) the network should distinguish between is denoted as \(N\)-way.

Next, we define the training procedure for optimization-based meta-learning, which this paper is focused on. It is defined in 2 steps:
\begin{enumerate*}[label={\arabic*)}]
  \item adaptation step, which computes adaptation weights in a form of a function \(\theta'(\theta)\), that minimizes task-specific error \(\mathcal{L}(y_s, \Phi(\theta', X_S))\);
  \item meta-gradient update, which updates the meta-weights \(\theta\).
\end{enumerate*}
The idea behind such training procedure is that by finding good weights \(\theta\), it will be possible to adapt to new previously unseen tasks with few training examples in the adaptation procedure. For classification, the loss function used is typically cross-entropy:
\begin{equation}
  \label{eq:loss-function}
  \mathcal{L}(y,\Phi(\theta, X)) = - \sum_i {y_i \log{\Phi(\theta, X_i)}}.
\end{equation}

\section{Literature Overview}

The meta-learning approaches are mainly divided into 3 broad categories~\cite{LearningToLearnFastBlog}: metric-based, model-based and optimization-based. Representatives of each group differ in the neural network design and training procedure. Applications exist in literally every field of machine learning~\cite{MAML,MetaLearningNlpSurvey,WangSurvey2020,MetaLearningFaceRecognitionGuo2017}, such as NLP, Reinforcement Learning, Face Verification, etc. In this work we focus on Image Classification.

Typically meta-learning methods are split into the following categories:
\begin{enumerate}
  \item Metric-based methods, where the goal is to define a neural network architecture that produces an embedding into a metric space and a similarity measure (metric). The distance between embeddings of the same class should be smaller than that of the different classes. Examples of such methods include Siamese Networks~\cite{SiameseNetworks}, Matching Networks~\cite{MatchingNetworks}, Prototypical Networks~\cite{PrototypicalNetworks}.
  \item In model-based methods the network architecture is designed, so that the model has explicit memory cells, which help the network to adapt quickly. Memory-Augmented Neural Networks~\cite{MetaLearningMemoryAugmentedNN} might serve as an illustrative example of this approach.
  \item In contrast to previous categories, optimization-based learning does not involve network architecture change, which means that conventional architectures for image classification can be used.
\end{enumerate}

One of the quintessential optimization-based methods is MAML~\cite{MAML}. It defines the training procedure as a 2\textsuperscript{nd}-order optimization problem. The method applicability has been shown in regression, classification and reinforcement learning. Two popular datasets were considered for image classification: Omniglot~\cite{Omniglot} and miniImageNet~\cite{MiniImageNetRavi2017,MatchingNetworks}, where MAML has beaten many of the previous methods with a margin. After MAML has been introduced, a lot of works have proposed its modifications. Reptile~\cite{Reptile} has simplified MAML training scheme, MAML++~\cite{HowToTrainMAML} has given practical recommendations on improving MAML training stability. In has been noted that while MAML++ has introduced more parameters to the network, total training time has decreased thanks to the performance optimizations proposed. Authors of Meta-SGD~\cite{Meta-SGD} note that by learning not only network weights, but also separate update coefficients for each of the weights, it is possible to achieve higher accuracy. However, the network training time and memory consumption has significantly increased as twice the number of the parameters should be optimized.

In contrast to previous works, in this paper we focus on improving the network adaptation and not training time. We assume that after the initial training, the network can be adapted to multiple tasks in an online format. Thus, minimizing adaptation time is an important problem. The results obtained in the paper will be applicable to many of the optimization-based algorithms, including but not limited to the ones mentioned above.

\section{Materials and Methods}\label{sec:materials}

In this work we propose a modification to the MAML algorithm. This class of algorithms is defined in terms of adaptation and meta-gradient update phases. The algorithm starts by randomly sampling a training task \(\mathcal{T}_i \sim \rho(\mathcal{T})\). To sample a task \(\mathcal{T}_i\) means to 1) randomly select \(N\) classes from all classes that are available in the dataset split (training, validation or test); 2) randomly select \(K\) images per each of the \(N\) classes for the Support Set and \(K_Q\) images for the Query Set. The first phase of the algorithm is adaptation, where MAML minimizes loss function~\cref{eq:loss-function} on the Support Set by performing several stochastic gradient descent steps. The algorithm iteratively builds model weights \(\theta_i^{(j)}(\theta)\) via \cref{eq:theta-adaptation}, note that \(\theta_i^{(0)}(\theta) \equiv \theta\):
\begin{equation}
  \label{eq:theta-adaptation}
  \theta_i^{(j)} = \theta_i^{(j - 1)} - \alpha \nabla_{\theta_i} \mathcal{L} \left (y_{S_i}, \Phi \left (\theta_i^{(j-1)}, X_{S_i} \right ) \right ).
\end{equation}
Having iteratively built the task specific weights \(\theta_i^{(j)}\), the algorithm updates the meta-weights \(\theta\) using \cref{eq:theta-meta-gradient-update}:
\begin{equation}
  \label{eq:theta-meta-gradient-update}
  \theta \leftarrow \theta - \beta \nabla_\theta \sum_{Q_i \in \mathcal{T}_i} {\mathcal{L} \left (y_{Q_i}, \Phi \left (\theta_i^{(P)}, X_{Q_i} \right ) \right )}.
\end{equation}

In essence, in \cref{eq:theta-meta-gradient-update} the algorithm updates the meta-weights \(\theta\) by averaging computed loss function \cref{eq:loss-function} on the Query Set for the neural network \(\Phi\) with weights \(\theta_i^{(P)}\) on several tasks \(\mathcal{T}_i\), \ie, in this step the algorithm backpropagates through the losses of all the task-specific adaptations. Throughout the paper we use 4 tasks for the meta-update step. Note, that in \cref{eq:theta-adaptation} task-specific weights \(\theta_i^{(j)}\) are computed on the Support Set, while in \cref{eq:theta-meta-gradient-update} Query Set is used for the loss computation. Also, in contrast to the conventional neural network training procedure the loss function is computed twice: first, to compute the adaptation weights \(\theta_i^{(P)}\) in \cref{eq:theta-adaptation}; second, to compute the resulting adaption loss in \cref{eq:theta-meta-gradient-update}. Additionally, in \cref{eq:theta-adaptation} the gradient is taken by task-specific weights \(\theta_i^{(j - 1)}\), and in \cref{eq:theta-meta-gradient-update} the gradient is taken by the meta-weights \(\theta\). Thus, as can be seen from \cref{eq:theta-adaptation,eq:theta-meta-gradient-update} the method requires Hessian computation during the meta-gradient update. Hence, this is a second-order optimization method. The whole training procedure is depicted in \cref{alg:maml-adaptation}. A more detailed description can be found in the original paper~\cite{MAML}.

\begin{algorithm}
  \caption{MAML adaptation procedure.}\label{alg:maml-adaptation}
  Randomly sample task \(\mathcal{T}_i\) from task space \(\rho(\mathcal{T})\)

  \For{each task \(\mathcal{T}_i = \{S_i, Q_i\}\), where \(S_i = {X_S, y_S}, Q_i = {X_Q, y_Q}\)}
  {
    \For{iteration \(j = 1 \ldots P\)}
    {
      Adapt the network via \cref{eq:theta-adaptation} using \(S_i\)
    }
  }
  Update meta-weights \(\theta\) via \cref{eq:theta-meta-gradient-update} using \(Q_i\) and the task specific weights \(\theta_i^{(P)}\).
\end{algorithm}

Next, we define our modified adaptation procedure. Given a convolutional neural network that has \(B\) layers, we define an adaptation pattern \cref{eq:lambda-pattern}, where \(\Lambda_j\) is an indicative function as defined in \cref{eq:lambda-l}, which indicates layers of the network that should be updated during backpropagation.

\begin{equation}
  \label{eq:lambda-pattern}
  \Lambda = \{\Lambda_1, \Lambda_2, \ldots, \Lambda_B\},
\end{equation}

\begin{equation}
  \label{eq:lambda-l}
  \forall l: \Lambda_l = 
  \begin{cases}
    1, & \text{if layer } l \text{ is updated},\\
    0, & \text{otherwise}.
  \end{cases}
\end{equation}

\begin{figure*}
  \centering
  \includegraphics[width=\linewidth]{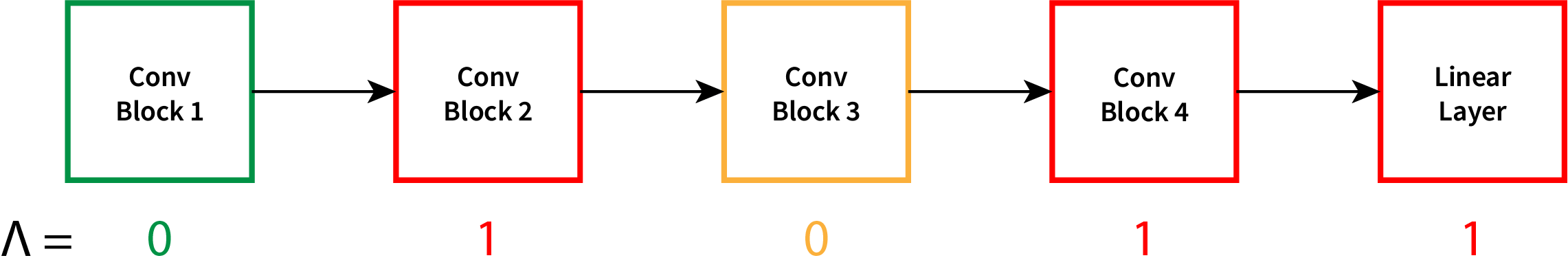}
  \caption{\(\Lambda\) pattern backpropagation scheme. Backpropagation is performed in order reverse to the arrows. In red -- gradients are computed, networks weights are updated; yellow -- gradients are computed, no network weight update; green -- both gradient computation and network weight update are skipped.}\label{fig:lambda-pattern-scheme}
\end{figure*}

We say that pattern is full if \(\forall l: \Lambda_l = 1\). In this case our adaptation phase will be equivalent to the one proposed in MAML. We consider all possible patterns \(\Lambda\), except \(\forall l: \Lambda_l = 0\), when no weights can be updated; thus, no adaptation is possible. We assume that updating only certain weights might be useful, because the neural networks tend to learn features that differ in complexity, the closer the layer is to the input the simple the features are~\cite{VisualizingNetworksZeiler2014}. Also, authors of Meta-SGD~\cite{Meta-SGD} have shown that by learning weight-specific learning rates the resulting quality was superior to the original MAML algorithm. However, Meta-SGD approach was much slower to train as both weights and learning rates have to be learned during the training procedure. Training time in our approach is intact. In contrast to previous works, we propose to update only certain weights; thus, essentially freezing some layers. This allows us to decrease gradient computations required during the adaptation phase as is shown in \cref{fig:lambda-pattern-scheme} for a convolutional network that contains 4 convolutional and a single fully connected (linear) layer (\(B = 5\)).

In \cref{fig:lambda-pattern-scheme} the backpropagation pass goes in the direction opposite to arrows. The architecture is taken as an example and can be arbitrary in practice. For the considered pattern \(\Lambda = \{0,1,0,1,1\}\):
\begin{enumerate}
  \item In the convolutional block~4 and the linear layers both the gradient is computed and the weights are updated.
  \item In convolutional block~3 the weights are not updated, but the gradients are computed as convolutional block~2 requires weight update.
  \item In convolutional block~1 no gradients computation or weight update are performed.
\end{enumerate}

Given the above-described \(\Lambda\) pattern description, the updated adaptation formula will look as follows:
\begin{equation}
  \label{eq:theta-lambda-pattern-adaptation}
  \theta_i^{(j)} = \theta_i^{(j - 1)} - \textcolor{red}{\Lambda} \alpha \nabla_{\theta_i} \mathcal{L} \left (y_{S_i}, \Phi \left (\theta_i^{(j-1)}, X_{S_i} \right ) \right ).
\end{equation}

\section{Experiments}

To conduct the experiments, we have reimplemented the MAML algorithm. The following paragraph describes the details.

The authors of MAML have defined a convolutional neural network architecture and have used it for miniImageNet experiments. This network is commonly referred to as ``CNN4'' in the literature. It has 4 convolutional blocks, followed by a linear layer. Each of the blocks has a convolutional layer with a kernel size of 3 and a padding of 1, followed by the Batch Normalization~\cite{BatchNorm}, ReLU activation and Max Pooling with a kernel size of 2. Number of filters in the convolutional layers is a configurable parameter. The authors have used 32, which we follow. Number of outputs in the linear layer is defined by \(K\) for \(K\)-way classification problem. Training is performed via Adam~\cite{Adam} gradient descent method as meta-optimizer with learning rate of \(\beta = 10^{-3}\), and \(\alpha = 0.01\) as the adaptation step size. Each model has been trained for 600 epochs. The authors used meta-batch size of 2 for 5-shot and 4 for 2-shot experiments. Instead, we consistently use meta-batch size of 4 as it leads to slightly better performance on CIFAR-FS~\cite{MetaLearningClosedFormSolvers} dataset during our experiments. Each epoch has 100 randomly sampled tasks \(\mathcal{T}_i\). For the gradient update \(N \cdot K\) samples are taken for \(N\)-shot \(K\)-way classification problem for training and 15 samples per class for evaluation, following~\cite{MiniImageNetRavi2017}.

In addition, we have modified the network adaptation procedure, so that it updates only weights defined by pattern \(\Lambda\) as defined in \cref{eq:lambda-pattern,eq:lambda-l,eq:theta-lambda-pattern-adaptation}.

For the experiments we have used the novel CIFAR-FS~\cite{MetaLearningClosedFormSolvers} dataset. It has been constructed from a well-known CIFAR-100~\cite{CIFAR-100} classification dataset. It has images of different kinds of mammals, reptiles, flowers, man-made things, \etc. The images are in color and have a size of \(32 \times 32\). In~\cite{MetaLearningClosedFormSolvers} it has been suggested to split 100 classes into train, validation and test sets. If it has been the non-few-shot neural network training, we would expect all of the 100 classes to be equally represented in each of the sets, only the images themselves would have been split. However, in case of few-shot learning different disjoint classes are taken into each subset. Specifically, 64, 16, and 20 classes have been selected for training, validation, and test set correspondingly. The exact classes that go into each split are important and are defined in~\cite{MetaLearningClosedFormSolvers}. By using different classes for training and testing, the adaptation to the new classes can be better estimated. After such training the model is expected to quickly adapt to the new unseen classes. We have taken the CIFAR-FS dataset for our experiments as it hasn't been analyzed by the MAML authors and is also faster to compute than miniImageNet.

All of the training procedures and time measurements were done on our own MAML implementation and tested on NVIDIA GTX 1050Ti GPU.

\section{Results}

Accuracies and timings for our MAML implementation on CIFAR-FS are presented in \cref{tab:results-cifar-fs}.

\begin{table}
  \caption{MAML accuracy and adaptation time on CIFAR-FS dataset.}\label{tab:results-cifar-fs}
  \centering
  \begin{tabular}{@{}lcccc@{}}
  \toprule
                            & 1-shot   & 5-shot   & 1-shot   & 5-shot  \\
                            & 2-way    & 2-way    & 5-way    & 5-way   \\
  \midrule
  Accuracy                  & 77.2~\%  & 87.6~\%  & 51.7~\%  & 70.3~\% \\
  Time                      & 38.43~ms & 40.70~ms & 41.67~ms & 45.35~ms
  \end{tabular} 
\end{table}

\begin{figure*}
  \centering
  \begin{subfigure}{0.49\linewidth}
    \centering\captionsetup{margin=2cm}
    \includegraphics[height=7cm]{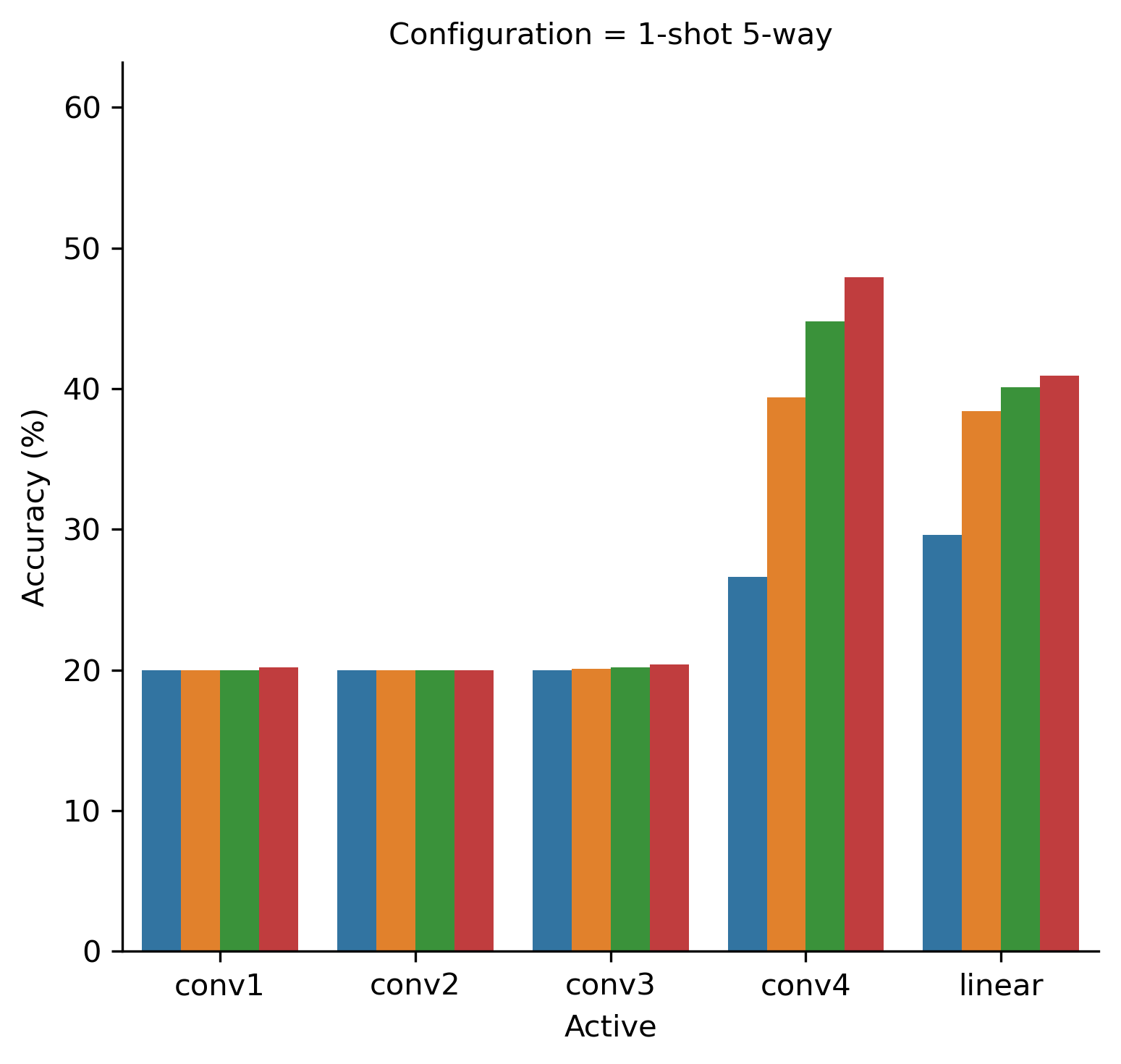}\caption{1-shot 5-way}\label{fig:adaptation-accuracy-1-shot-5-way}
  \end{subfigure}
  \begin{subfigure}{0.49\linewidth}
    \centering\captionsetup{margin=2.3cm,singlelinecheck=off,justification=raggedright}
    \includegraphics[height=7cm]{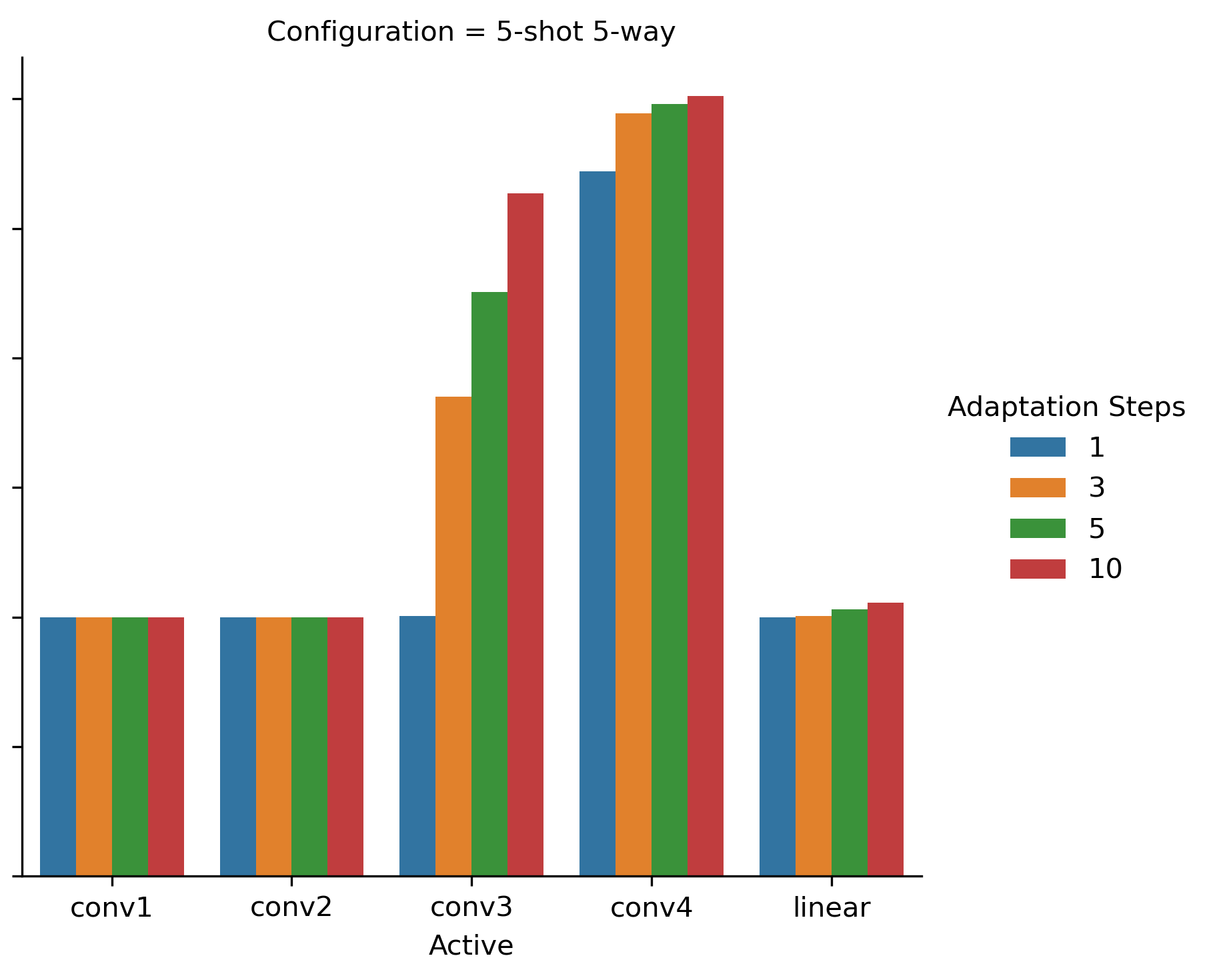}\caption{5-shot 5-way}\label{fig:adaptation-accuracy-5-shot-5-way}
  \end{subfigure}
  \caption{Adaptation accuracy for trivial \(\Lambda\) patterns, \ie, when only a single layer is updated during adaptation (5-way).}\label{fig:adaptation-accuracy-5-way}
\end{figure*}

In \cref{eq:theta-lambda-pattern-adaptation} we have proposed a modified adaptation formula, where only a part of weights is updated during the adaptation procedure. To begin with, let us consider only trivial patterns \(\Lambda\), where only one network layer is updated during the adaptation procedure. In \cref{fig:adaptation-accuracy-1-shot-5-way,fig:adaptation-accuracy-5-shot-5-way} we conduct the experiment for 1-shot 5-way and 5-shot 5-way configurations correspondingly. The results are presented on the test set. To see the impact of the number of adaptation steps, we show the accuracies for \(P = 10\) (default) and 1, 3, 5 adaptation steps. As it can be seen, the model accuracy differs significantly between the configurations.

For 1-shot 5-way, learning one of the first three convolutional layers only has no effect, the accuracy remains on the level of random guessing (20\%). However, training either convolutional layer 4 or the last linear layer improves the model accuracy. Note, that the number of parameters in layers differs. In \cref{tab:params-per-layer}, we show the number of parameters in each layer. Note, that the final layer has different number of parameters depending on \(N\) output classes. It can be seen, that the first convolutional layer and the final linear (fully connected) layers have fewer parameter than inner convolutional blocks.

For 5-shot 5-way scenario we see a different picture. Only convolutional layers 3 and 4 have positive impact on the performance if adapted alone. Interestingly, the number of adaptation steps has a significant impact on the performance with only convolutional layer \#3 enabled. As we will see later, such an impact is higher, than when all network layers are updated during the adaptation.

\begin{table}
  \caption{Number of parameters for each layer. Note, that the number of parameters is different in the final Linear layer depending on 2- or 5-way problem setting.}\label{tab:params-per-layer}
  \centering
  \begin{tabular}{lr}
  \toprule
  Layer Name    & \# Parameters \\
  \midrule
  Conv Block 1  &     960       \\
  Conv Block 2  &   9,312       \\
  Conv Block 3  &   9,312       \\
  Conv Block 4  &   9,312       \\
  Linear        &               \\
  \quad 2-way   &   1,602       \\
  \quad 5-way   &   4,005       \\
  Total         &               \\
  \quad 2-way   &  30,498       \\
  \quad 5-way   &  32,901       \\
  \end{tabular} 
\end{table}

\begin{figure*}
  \centering
  \begin{subfigure}{0.49\linewidth}
    \centering\captionsetup{margin=2cm}
    \includegraphics[height=7cm]{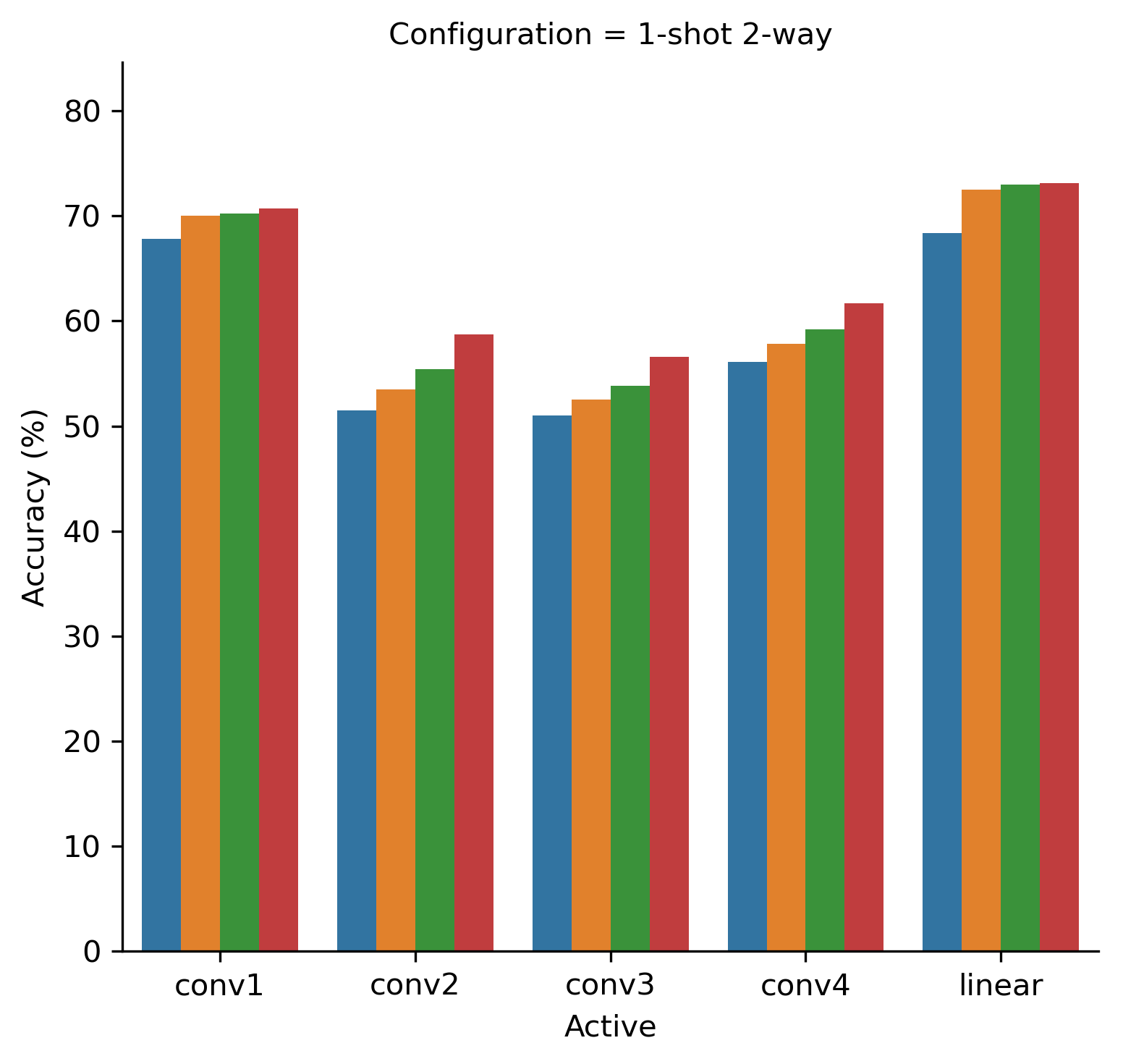}\caption{1-shot 2-way}\label{fig:adaptation-accuracy-1-shot-2-way}
  \end{subfigure}
  \begin{subfigure}{0.49\linewidth}
    \centering\captionsetup{margin=2.3cm,singlelinecheck=off,justification=raggedright}
    \includegraphics[height=7cm]{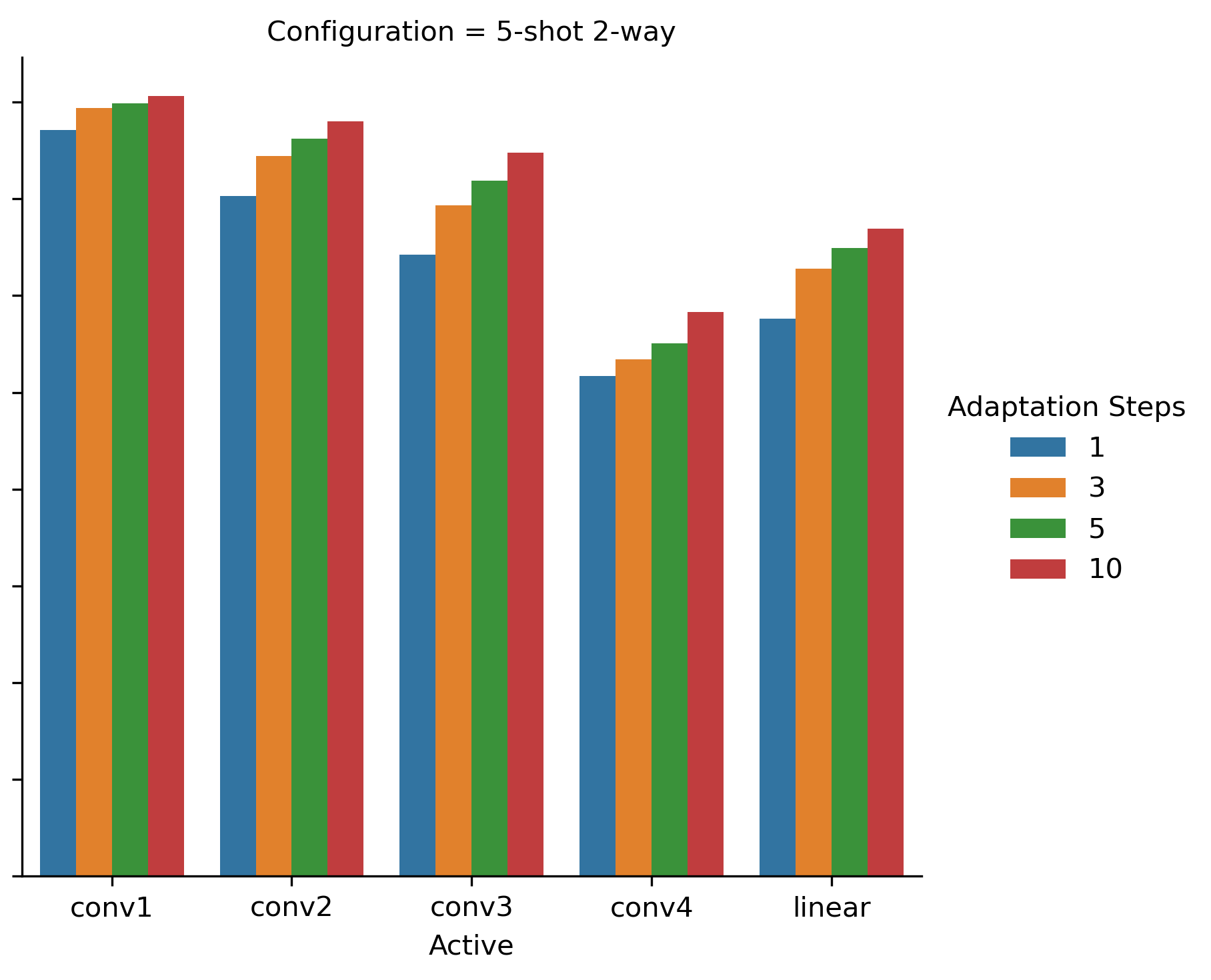}\caption{5-shot 2-way}\label{fig:adaptation-accuracy-5-shot-2-way}
  \end{subfigure}
  \caption{Adaptation accuracy for trivial \(\Lambda\) patterns, \ie, when only a single layer is updated during adaptation (2-way).}\label{fig:adaptation-accuracy-2-way}
\end{figure*}

In \cref{fig:adaptation-accuracy-1-shot-2-way,fig:adaptation-accuracy-5-shot-2-way} we depict a similar experiment for 1-shot 2-way and 5-shot 2-way configurations correspondingly. Note, that random guessing baseline for these configurations is now at 50\%, so the lower bound for accuracy is now higher than in \cref{fig:adaptation-accuracy-5-way}. Here we see an opposite trend, where updating the first layers also has a positive impact on the resulting accuracy. Contrasting to previous experiment, updating exclusively the convolutional block 4 doesn't provide the best results in either configuration.

\begin{figure*}
  \centering
  \begin{subfigure}{0.49\linewidth}
    \centering\captionsetup{margin=2cm}
    \includegraphics[height=7cm]{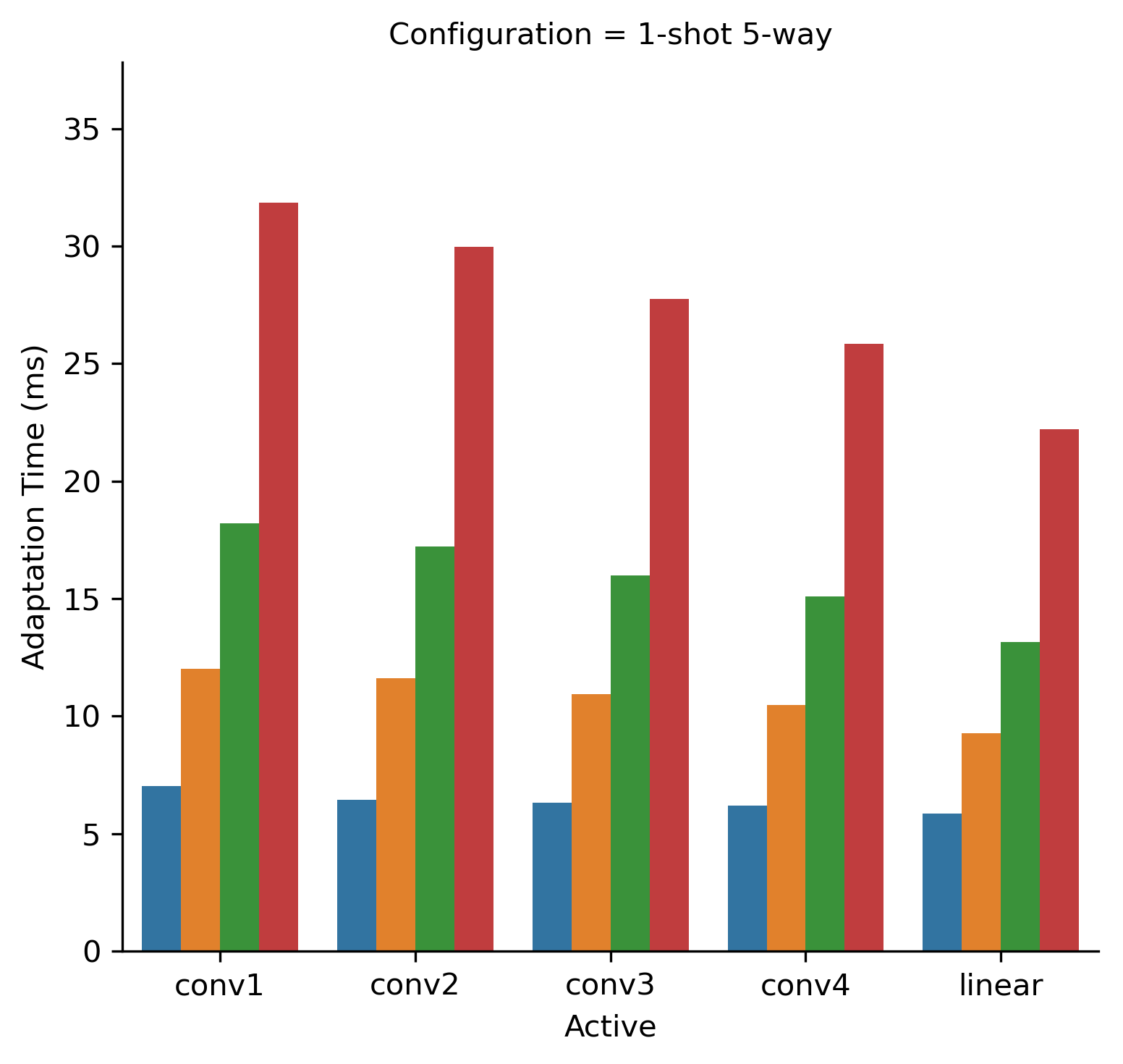}\caption{1-shot 5-way}\label{fig:adaptation-time-1-shot-5-way}
  \end{subfigure}
  \begin{subfigure}{0.49\linewidth}
    \centering\captionsetup{margin=2.3cm,singlelinecheck=off,justification=raggedright}
    \includegraphics[height=7cm]{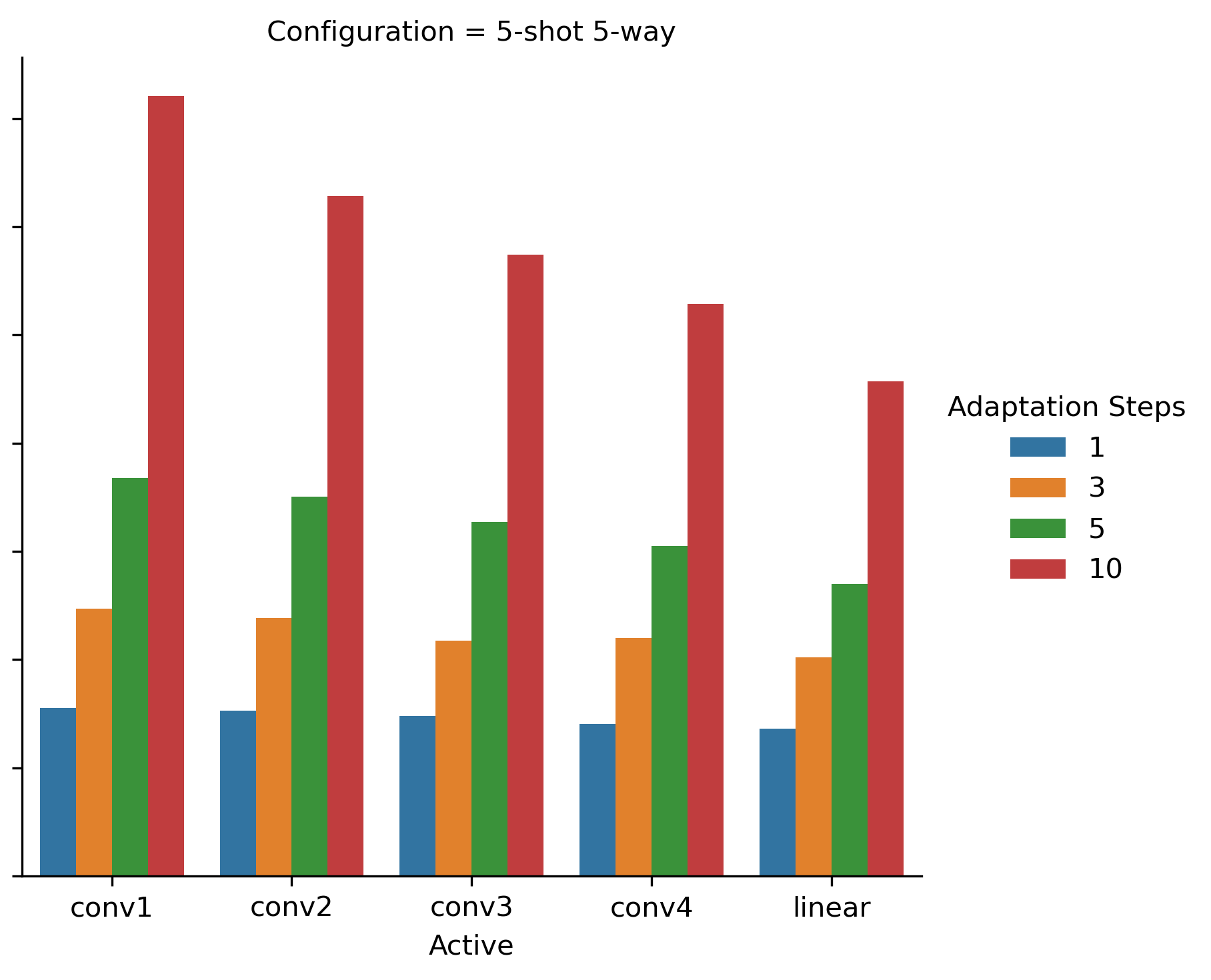}\caption{5-shot 5-way}\label{fig:adaptation-time-5-shot-5-way}
  \end{subfigure}
  \caption{Adaptation time for trivial \(\Lambda\) patterns (5-way).}\label{fig:adaptation-time-5-way}
\end{figure*}

In \cref{fig:adaptation-time-5-way} we show adaptation time for 1-shot 5-way and 5-shot 5-way configurations for all trivial \(\Lambda\) patterns. As can be seen, selected \(\Lambda\) pattern and the number of adaptation steps have a significant impact on the adaptation speed. A similar trend is observed in 2-way configurations.

\begin{figure}
  \centering
  \includegraphics[width=\linewidth]{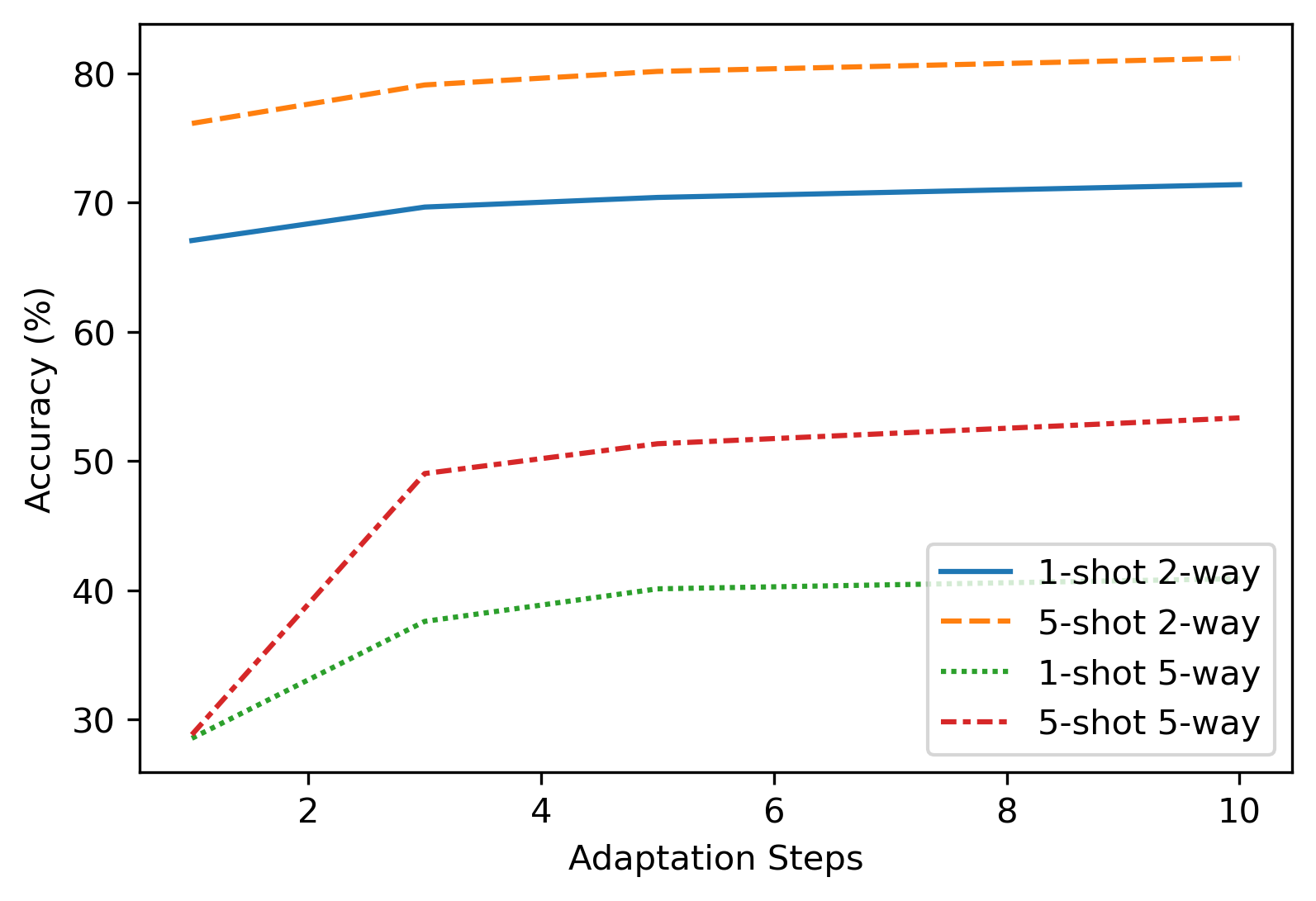}
  \caption{Accuracy averaged for all \(\Lambda\) patterns for different \(N\)-shot \(K\)-way problems with respect to the number of adaptation steps \(P\).}\label{fig:average-accuracy}
\end{figure}

\begin{figure}
  \centering
  \includegraphics[width=\linewidth]{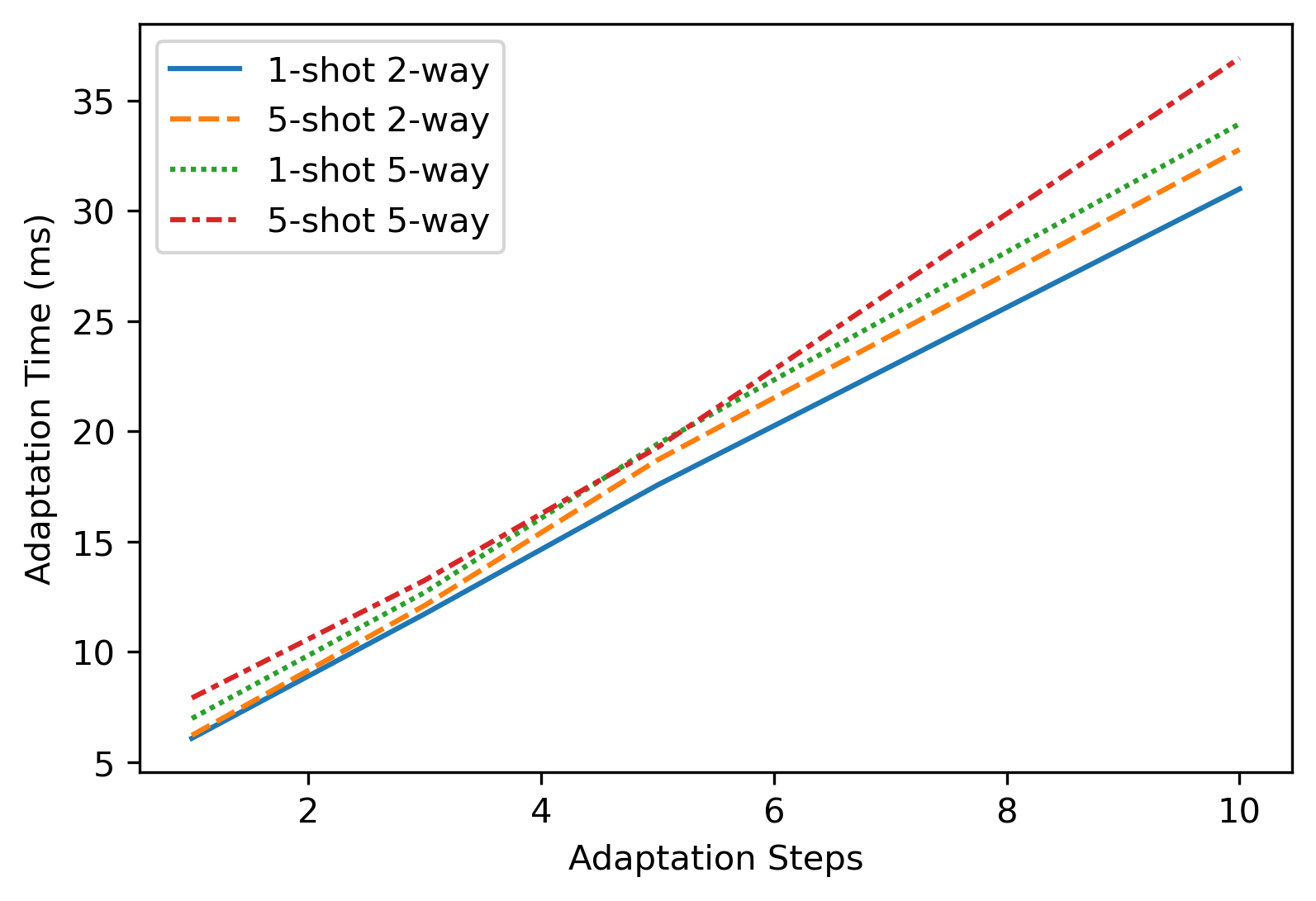}
  \caption{Adaptation time averaged for all \(\Lambda\) patterns for different \(N\)-shot \(K\)-way problems with respect to the number of adaptation steps \(P\).}\label{fig:average-adaptation-time}
\end{figure}

In \cref{fig:average-accuracy} we show the model accuracy for each of the four scenarios and in \cref{fig:average-adaptation-time} we depict the corresponding timings, both shown with respect to the number of the adaptation steps. As before, the experiments have been conducted for \(P = 1, 3, 5\) and 10 adaptation steps. The results between those reference points have been linearly interpolated. The presented accuracies and timings are averaged for all 31 possible \(\Lambda\) patterns. Note, that throughout the article we exclude pattern \(\forall l: \Lambda_l = 0\), as no weights can be changed for such pattern, therefore no adaptation is possible. As can be seen, while the adaptation time grows linearly with the number of adaptation steps, the accuracy growth plateaus at around 5 adaptation steps. Actually, for the full pattern \(\Lambda\) increasing the number of adaptation steps from 5 to 10 has less than 0.3\% improvement in accuracy. In typical practical scenarios such an improvement is insignificant. Thus, we suggest that performing 10 adaptation steps is redundant.

Next, we try to search for such a pattern \(\Lambda\) and number of adaptation steps, so that the resulting accuracy drops no more than 0.07 times the full pattern accuracy. We see such a quality degradation threshold reasonable for practical applications. Obviously, the approach we propose can be applied with an arbitrary quality degradation threshold. We search for such patterns and show them in \cref{tab:lambda-pattern-metrics-cifar-fs}. Based on this table, we suggest using the \(\Lambda^* = \{1,0,1,1,1\}\), which offers a factor of 3.0 speed improvement with an insignificant quality loss. It can be seen that pattern \(\Lambda = \{0,1,1,1,1\}\) also suits the specified criteria and has a slightly higher (factor of 3.1) performance improvement. However, it has a significantly lower performance for both of the 2-way configurations, degrading on 2.5\% and 3.2\% relative to the best selected pattern \(\Lambda^*\). We consider such a degradation not worth the additional speed up. The fact that enabling first CNN layer is significant for the 2-way learning accuracy, closely follows the presented above description of \cref{fig:adaptation-accuracy-2-way}. Also, not to be mistaken, in \cref{fig:adaptation-accuracy-5-way,fig:adaptation-accuracy-2-way,fig:adaptation-time-5-way}, we had only one layer updated during the adaptation phase (thus \(\sum_l \Lambda_l = 1\)). However, the best selected pattern \(\Lambda^*\) has all expect one layer updated.

\begin{table*}
  \caption{Adaptation speedup depending on pattern \(\Lambda\) and the number of adaptation steps. Patterns with loss degradation of less than 7\% (relative to the full pattern \(\Lambda\) and 10 adaptation steps) are shown.}\label{tab:lambda-pattern-metrics-cifar-fs}
  \centering
  \begin{tabular}{@{}cccccccc@{}}
    \toprule
    Adaptation &  Pattern   & 1-shot 2-way & 5-shot 2-way & 1-shot 5-way & 5-shot 5-way & Mean Adaptation &  Relative \\
    Steps      &            &      (\%)    &      (\%)    &      (\%)    &      (\%)    &    Time (ms)    &  Speedup  \\
    \midrule
     3         &  0,1,1,1,1 &         74.7 &         83.2 &         49.3 &         69.7 &            13.3 &       3.1 \\
     3         &  1,0,1,1,1 &         76.6 &         85.9 &         49.3 &         69.8 &            13.9 &       3.0 \\
     3         &  1,1,1,1,1 &         76.6 &         87.2 &         49.3 &         70.0 &            15.0 &       2.8 \\
     5         &  0,1,1,1,1 &         75.2 &         83.9 &         51.5 &         69.9 &            20.0 &       2.1 \\
     5         &  1,0,1,1,1 &         76.9 &         86.2 &         51.4 &         70.1 &            21.1 &       2.0 \\
     5         &  1,1,1,1,1 &         77.0 &         87.4 &         51.6 &         70.2 &            22.6 &       1.8 \\
    10         &  0,1,1,1,1 &         75.4 &         84.6 &         51.7 &         70.1 &            36.1 &       1.2 \\
    10         &  1,0,1,1,1 &         77.1 &         86.6 &         51.7 &         70.1 &            38.6 &       1.1 \\
    10         &  1,1,1,1,1 &         77.2 &         87.6 &         51.7 &         70.3 &            41.5 &       1.0 \\
    \bottomrule
  \end{tabular}
\end{table*}

Finally, we pose a question, whether updating only a part of the neural network weights can improve the method accuracy. It turns out, that in an extreme case of learning with a single adaptation step (\(P = 1\)), significant improvement in 5-way adaptation performance is achieved by updating with a partial pattern \(\Lambda\). This is shown in \cref{tab:single-gradient-step-accuracy}.

\begin{table}
  \caption{Accuracy improvement for \(P = 1\) gradient step adaptation with pattern selection.}\label{tab:single-gradient-step-accuracy}
  \centering
  \begin{tabular}{@{}lcccc@{}}
  \toprule
  \multirow{2}{*}{Pattern}    & 1-shot   & 5-shot   & 1-shot            & 5-shot          \\
                              & 2-way    & 2-way    & 5-way             & 5-way           \\
  \midrule
  \(\Lambda = \{1,1,1,1,1\}\) & 74.3~\%  & 86.0~\%  & 36.8~\%           & 20.4~\%         \\
  \(\Lambda = \{1,1,0,1,1\}\) & 74.3~\%  & 83.1~\%  & \textbf{36.9~\%}  & \textbf{53.1~\%}
  \end{tabular} 
\end{table}

We have also performed a search of all cases, when our approach gives better results than the original with \(P = 1\). The results are shown in \cref{tab:single-gradient-step-pattern-selection}.

\begin{table*}
  \caption{Accuracy improvement for \(P = 1\) gradient step adaptation with pattern selection if the pattern is selected per configuration.}\label{tab:single-gradient-step-pattern-selection}
  \centering
  \begin{tabular}{@{}lcccc@{}}
  \toprule
                                          & 1-shot 2-way     & 5-shot 2-way     & 1-shot 5-way     & 5-shot 5-way    \\
  \midrule
  Accuracy on \(\Lambda = \{1,1,1,1,1\}\) & 74.3~\%          & 86.0~\%          & 36.8~\%          & 20.4~\%          \\
  Accuracy on selected \(\Lambda\)        & \textbf{74.5~\%} & \textbf{86.2~\%} & \textbf{36.9~\%} & \textbf{54.8~\%} \\
  Selected Pattern \(\Lambda\)            & 1,1,1,0,1        & 1,1,1,0,1        & 1,1,0,1,1        & 1,1,0,1,0
  \end{tabular} 
\end{table*}

\section{Discussion}

In~\cite{VisualizingNetworksZeiler2014} it has been shown that each trained neural network's convolutional layer has a different meaning. The first layer learns to detect simple features, like edges, lines or color gradients. The second layer increases the complexity and understands simple shapes, \eg, circles, corners or stripes, while the last layers learn high-level features, such as eyes, faces, text-like objects, \etc. The exact features learned, obviously, depend on the training dataset, still such logic is retained. In the few-shot classification scenario the tasks differ by the types of objects that the model has to classify. As we have described in the experiments section, train and test sets have disjoint classes included. Thus, it might be reasonable to expect that only the last layers of the network should be changed to adapt to the new tasks and classes. This is exactly what we see in the case of 5-way classification as is shown in \cref{fig:adaptation-accuracy-5-way}. However, such a statement contradicts to the experiment results from \cref{fig:adaptation-accuracy-2-way}.

To understand the contradiction, we examine the original CIFAR-100 dataset. There image labels (classes) are structured to form larger coarse groups. For instance, coarse class (or superclass) ``aquatic mammals'' contains ``beaver'', ``dolphin'', ``otter'', ``seal'', ``whale''. Other examples of superclasses include ``fish'', ``large carnivores'', ``household electrical devices'', \etc. From the examples we have picked, it becomes obvious that images from different classes have significantly different color gamut. Images of aquatic mammals and fish typically contain blue and gray colors, while large carnivores might have more yellow and green. In case of 2-way classification it is more probable that the network has to classify only between subclasses of the same superclass. Consequently, we assume that if the first layer is updated in a 2-way few-shot learning scenario, the new weights better adjust to features with different color gamut. We see this as an analogy to how a human eye works: it adjusts the amount of light coming to the retina by expanding or contracting the pupil, so that it becomes easier to see the details.

From \cref{tab:lambda-pattern-metrics-cifar-fs} we see that keeping the inner layers stale is the most fruitful way to improve the performance, with little to no quality loss. A substantial increase in adaptation speed has been achieved with a target quality loss set to 7~\% relative to the original pattern \(\Lambda = \{1,1,1,1,1\}\) and \(P = 10\) adaptation steps. The actual quality loss turns out to be even smaller as we have skipped slightly faster, but worse pattern \(\Lambda = \{0,1,1,1,1\}\). Thereby, with the best \(\Lambda^* = \{1,0,1,1,1\}\) and \(P = 3\) adaptation steps, we achieve a factor of 3.0 speed improvement. Our quality losses are the following: 1-shot 2-way is 0.78~\%, 5-shot 2-way is 1.97~\% 1-shot 5-way is 4.86~\% and 5-shot 5-way is 0.71~\%. Even smaller quality losses can be achieved by consulting \cref{tab:lambda-pattern-metrics-cifar-fs}. Note, that these are relative quality losses. If the losses are computed in absolute terms, they are even more negligible. Thus, we have achieved a significant adaptation time reduction with small-enough quality loss.

We also discuss a way to improve algorithm quality by selecting a pattern \(\Lambda\). In an extreme case of a single adaptation step, simply avoiding the inner layer update has helped to improve the overall model quality as is shown in \cref{tab:single-gradient-step-accuracy}. In addition, we have been able to find such a pattern \(\Lambda\) for each of the few-shot learning configurations, such that it improves the model performance for \(P = 1\) adaptation step in \cref{tab:single-gradient-step-pattern-selection}. It is curious that no such behavior is observed in cases when \(P > 1\). To the best of our knowledge such behavior has not been previously observed and should be further investigated.

\section{Conclusion}

MAML is an optimization-based few-shot learning method that is able to learn by using only a few samples per class. Many algorithms follow the learning scheme proposed in MAML. In this work we solve its problems of
\begin{enumerate*}[label={\arabic*)}]
  \item long adaptation time, and
  \item poor performance in cases when a single adaptation step is used.
\end{enumerate*}

In this work \(\Lambda\) pattern method has been introduced. This method reduces the number of gradient computations in MAML adaptation phase. By selecting an appropriate adaptation pattern, we have significantly improved the method in the following areas:
\begin{enumerate*}[label={\arabic*)}]
  \item long MAML adaptation time has been decreased by the factor 3 with minimal accuracy loss;
  \item accuracy in cases when only a single adaptation step is used has been substantially improved.
\end{enumerate*}

The improvement of adaptation time of the widespread MAML algorithm will enable its applicability on less powerful devices and will in general decrease the time needed for the algorithm to adapt to new tasks.

Prospects for further research are to investigate a way of a more robust automatic pattern selection scheme for an arbitrary training dataset and network configuration.

\begin{small}
  \subsection*{Funding}

  The work is supported by the state budget scientific research project of Dnipro University of Technology ``Development of New Mobile Information Technologies for Person Identification and Object Classification in the Surrounding Environment'' (state registration number 0121U109787).

  \printbibliography

\end{small}

\end{document}